\documentclass[9pt,twocolumn,twoside]{osajnl}

\journal{ao} 

\setboolean{shortarticle}{false} 

\ifthenelse{\boolean{shortarticle}}{\colorlet{color2}{color2b}}{\colorlet{color2}{color2}} 

\usepackage{subcaption}
\usepackage{multirow}
\usepackage{amsmath}

\DeclareMathOperator*{\argmin}{argmin} 
\newcommand\Illum[1]{E_{#1}}
\newcommand\Reflect[1]{r_{#1}}
\newcommand\Surf[1]{\mathcal{S}_{#1}}
\newcommand\Point[1]{P_{#1}}

\newcommand\Normal[1]{\vec{N_{#1}}}
\newcommand\Dsurf[1]{dP_{#1}}

\newcommand\rh[2]{\rho^{#1}_{#2}}

\DeclareMathOperator{\diag}{diag}
\DeclareMathOperator{\vect}{vec}

\newcounter{append}

\title{Spectral reflectance estimation from one RGB image using self-interreflections in a concave object }

\author[1,*]{Rada Deeb}
\author[1]{Damien Muselet}
\author[1]{Mathieu Hebert}
 \author[1]{Alain Trémeau}

\affil[1]{University of Lyon, UJM-Saint-Etienne, CNRS, Institut d'Optique Graduate School, Laboratoire Hubert Curien UMR 5516, F-42023, SAINT-ETIENNE, France}

\affil[*]{Corresponding author: rada.deeb@univ-st-etienne.fr}

\dates{Compiled \today}

\ociscodes{ (110.4234) Multispectral imaging; (100.3190) Inverse problems;  (330.1710) Color, measurement}

\doi{\url{http://dx.doi.org/10.1364/ao.XX.XXXXXX}}

\begin{abstract}
Light interreflections occurring in a concave object generate a color gradient which is characteristic  of the object's spectral reflectance. In this paper, we use this property in order to estimate the spectral reflectance of matte, uniformly colored, V-shaped surfaces from a single RGB image taken under directional lighting.    
 First, simulations show that using one image of the concave object is equivalent to, and can even outperform, the state of the art approaches based on three images taken under three lightings with different colors.  Experiments on real images of folded papers were performed  under unmeasured direct sunlight. The results show that our interreflection-based approach outperforms existing approaches even when the latter are improved by a calibration step. 
The mathematical solution for the interreflection equation and the effect of surface parameters on the performance of the method are also discussed in this paper.
\end{abstract}

\setboolean{displaycopyright}{true}

\begin{document}

\maketitle
\thispagestyle{fancy}

\ifthenelse{\boolean{shortarticle}}{\ifthenelse{\boolean{singlecolumn}}{\abscontentformatted}{\abscontent}}{}


\section{Introduction} 

Can spectral reflectance of surfaces be measured with RGB cameras? This question is of high interest in many application fields since cameras are affordable and widely spread-used devices, in contrast to spectrophotometers or hyperspectral cameras. 
In computer vision, spectral description is often needed instead of tristimulus values.  Tri-values, such as CIE 1931 XYZ or CIELAB, may not be sufficient in applications such as paint selection, cosmetics industry, fruit quality assessment, telemedicine and eHeritage. These applications require spectral acquisition of the light signal reflected by the object in order to allow the user to detect and avoid the case of metamerism (the case where different spectral reflectances map to the same tri-value). In order to obtain spectral data, there exist two main methods: the use of spectral cameras, or the use of RGB images captured under different light sources. Spectral cameras give very accurate results, but they are expensive, and therefore not affordable for the broadest range of customers. As an alternative to these spectral cameras,  various approaches have been proposed using directly RGB images to extract spectral data. 

The state of the art in spectral reflectance estimation of surfaces from RGB images can be divided into two categories: \textit{direct methods} as called in \cite{zhao07,ribes08}, also named \textit{observational-model based methods} in \cite{heikkinen16}, and \textit{indirect methods} as called in \cite{zhao07,ribes08}, also named \textit{learning-based methods} in \cite{heikkinen16}. In direct methods, the Spectral response of sensors and the spectral power distribution (SPD) of illuminants are considered to be known. A common approach of these methods is to combine trichromatic imaging with multiple light sources \cite{park07,heikkinen08,chi10,fu16,han14,jiang12,khan13}. Park et al. \cite{park07} obtained spectral information by using RGB camera with a cluster of light sources with different spectral power distributions. They model the spectral reflectance of surfaces with a set of basis functions in order to reduce the dimensionality of the spectral space \cite{dannemiller92}. However, this approach received little attention mainly because of the need of several illuminants, which is incompatible with common situations where the surface is illuminated by ambient light. Later, Jiang et al. \cite{jiang12} proposed a more convenient solution, still based on various illuminants, by using commonly available lighting conditions, such as daylight at different times of a day, camera flash, ambient, fluorescent and tungsten lights.  However, a calibration step is required in order to handle variations in color temperature and spectral
properties of the used illuminants. More recently, Khan et al. \cite{khan13} proposed the use of a portable screen-based lighting setup in order to estimate the spectral reflectance  of the considered surface. The portable screen was used to give three lightings with red, green and blue colors. The capturing device is an RGB camera, therefore based on three different spectral responses, the spectral reflectance of the surface is expressed by nine coordinates in a basis of nine spectra. These basis functions are obtained by eigendecomposition of spectral reflectances of $1257$ Munsell color chips discarding the part of the spectra that corresponds to the sensor-illuminant null-space. The use of several light sources is in general mandatory to increase the number of equations needed to get the spectral values, together with the use of basis functions in order to decrease the dimensionality of the problem, therefore the number of spectral values to estimate.  

On the other hand, leaning-based approaches do not require any pre-knowledge of spectral responses of the sensors or spectral power distribution of the lighting system. They can be used with a single light source \cite{heikkinen07,heikkinen08,heikkinen13,heikkinen16}. However, these approaches depend on the quality of the learning set and on the choice of the regression method. It has been shown in \cite{heikkinen16}, that if no high quality learning set is available, using multiple light sources becomes necessary in order to improve the quality of the results.

In this work, we focus on relaxing the condition of multiple lightings and the need of calibrated settings in observational-based approaches by using a concave surface rather than a convex one. This allows us to take benefit of the light interreflection phenomenon which takes place in every material concavity. This phenomenon provokes color gradients in the image of the concave surface characteristics of the material's spectral reflectance. The latter can be therefore retrieved in a concrete application by using only one RGB image taken under daylight without the necessity of calibrating the camera under the acquisition configurations or measuring the incident irradiance.

\begin{figure}
	\centering
	\begin{subfigure}[b]{\textwidth/9}
		\includegraphics[width=\textwidth]{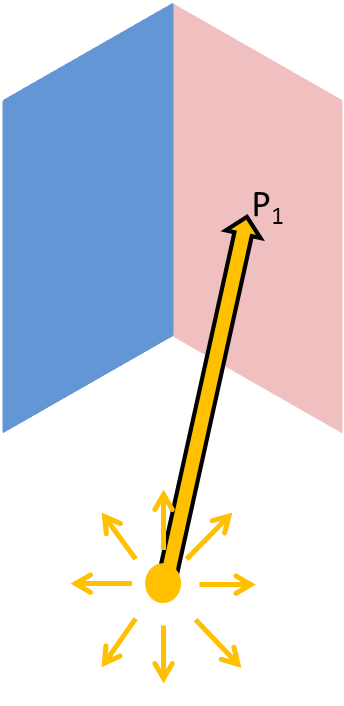}
		\caption{}
		\label{fig:1bounces} 
	\end{subfigure} \hfill
	\begin{subfigure}[b]{\textwidth/9}
		\includegraphics[width=\textwidth]{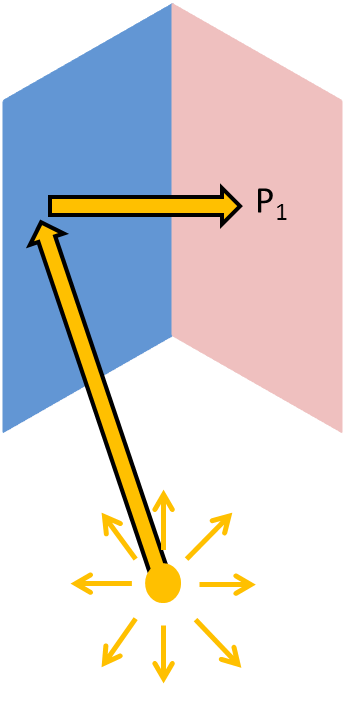}
		\caption{}
		\label{fig:2bounces}
	\end{subfigure} \hfill
	\begin{subfigure}[b]{\textwidth/9}
		\includegraphics[width=\textwidth]{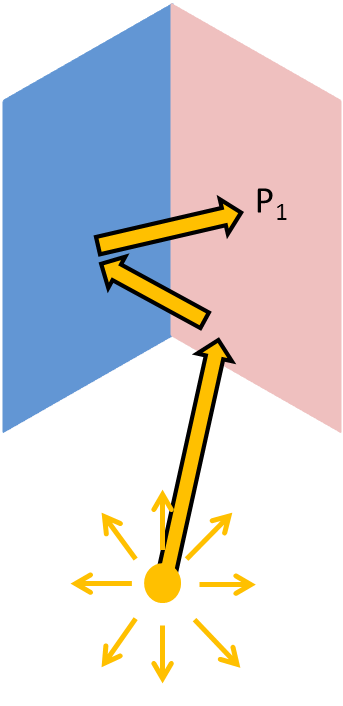}
		\caption{}
		\label{fig:3bounces}
	\end{subfigure}	
	\caption{Decomposition of the irradiance received by $\Point{1}$: (a) Direct light, (b) first bounce of interreflection corresponding to what we call a two-bounce model, (c) second bounce of interreflection. For the sake of clarity, the sums over all the pixels (multiple arrows) are not illustrated here.}
	\label{fig:bounces}
\end{figure}

Interreflections denote the fact that all points in a concave surface mutually illuminate each other during a multiple light reflection process, depending on the reflective properties and the geometrical shape of the surface. The simplest case is a flat diffusing surface bent into two flat panels, called a V-shaped surface and illustrated in Figure (\ref{fig:bounces}) where the first successive light bounces on the two panels are featured. The phenomenon of interreflections has gained attention in computer graphics in order to improve the rendering quality of scenes where various objects exchange light between each other (typically a colored object close to a white wall) \cite{byrd99, pharr16, jensen01}. This phenomenon has also been studied in the domain of computer vision, mainly in order to remove this effect from the images to be able to retrieve the shape of an object (shape-from-shading methods) \cite{Nayar91,Nayar92,Seitz05,Funt93,Liao11,fu14}. However, some approaches \cite{Funt91,drew90,ho90} in the literature focused on the use of interreflections as extra information in order to obtain surface spectral reflectance and illuminant spectral power distribution. 
The used methods were based on adjacent panels having different spectral reflectances, and only one bounce of interreflected light was considered. It has also been shown by Chandraker et al. \cite{chandraker05} that interreflections can resolve generalized bas-relief ambiguity in uncalibrated photometric stereo as interreflections are distance-dependent phenomena. In a recent work	\cite{deeb17}, the importance of using infinite bounces instead of two bounces of light was shown and a model of interreflection for Lambertian surfaces taking into consideration infinite bounces was proposed.

Starting form the same observation as Funt et al. \cite{Funt91}, we think that a scene with interreflections holds a lot of information about the physical properties that lie behind. However, instead of using only one bounce of interreflection, thus only one RGB color representing interreflections, we propose to take into consideration all light bounces and to model the variations of RGB values per surface element by using the model proposed in \cite{deeb17}. These variations can be seen as a crucial source of information for spectral reflectance estimation.

In this work, we first introduce the used interreflection model \cite{deeb17} which relates the variations of RGB values over a Lambertian concave surface to the reflectance of the surface, its shape, the illuminant and the used camera. Then, this model is used for spectral reflectance estimation applied in the  case of a folded surface with a uniform spectral reflectance. The main question we try to answer is, having a matte paper for example, how accurate its spectral reflectance can be estimated from a single RGB image by simply folding it? In other words, how much spectral information can interreflections carry in the RGB image?

The paper is structured as follows: we first present the infinite-bounce interreflection model in Section \ref{interModel} and the spectral reflectance estimation method in Section \ref{spectEstimation}. We then report our experimental results in Section \ref{results} and propose a discussion regarding the performance of the method in Section \ref{discussion}. Finally, conclusions are shown in Section \ref{conclusion}. 
 
\section{Interreflection model} \label{interModel}

In this section, we explain the used interreflection model for Lambertian surfaces \cite{deeb17}, starting by the bases of this model and ending by its extension to a spectral form.

The total irradiance,  $\Illum{}$, in every point of a concave surface is the sum of direct irradiance, received from the light source, and indirect irradiance, made of a collection of light rays reflected once, twice, three times and so on, from each point to each other point of the surface before reaching the considered point. For Lambertian surfaces, irradiance in a given point $\Point{i}$, received after one bounce of light from all other points $\Point{j}$ of the surface with respective reflectances $\Reflect{j}$, can be related to the radiance after one bounce of light,$L_1$, or to the irradiance received form direct light, $\Illum{0}$, as follows:

\begin{align} \label{eq:E1Int}
\Illum{1}(\Point{i}) &= \int_{\Point{j} \in \Surf{}} L_1 K(\Point{j}, \Point{i})\Dsurf{j} \nonumber \\
&= \int_{\Point{j} \in \Surf{}} \Reflect{j} \frac{\Illum{0}}{\pi} K(\Point{i}, \Point{j})\Dsurf{j}, 
\end{align}
where $\Dsurf{j}$ denotes an infinitesimal area around point $\Point{j}$, and the function $K$, called geometrical kernel, is defined for every pair of points in terms of the euclidean distance, $\Point{i}\Point{j}$, between them, the surface normal of each of them, and a visibility term. 
For a pair of points $\Point{i}$ and $\Point{j}$ with surface normals, $N_i$ and $N_j$ respectively, geometrical kernel is given by:

\begin{equation}  \label{eq:kernel}
K(\Point{i},\Point{j}) =  \frac{(\Normal{i}.\vec{\Point{i}\Point{j}})(\Normal{j}.\vec{\Point{j}\Point{i}})V(\Point{i},\Point{j}) }{\Point{i}\Point{j}^4},
\end{equation}
where  $V(\Point{i},\Point{j})$ is the visibility term which takes $1$ if the areas around these points can see each other and 0 otherwise.

Similarly, taking into account the rays reflected twice on every pair of points with respective reflectances $\Reflect{j}$ and $\Reflect{j'}$:

\begin{equation} \label{eq:E2Int}
\Illum{2}(\Point{i}) = \int_{\Point{j} \in \Surf{}}\int_{\Point{j'} \in \Surf{}} \Reflect{j} \Reflect{j'} \frac{\Illum{0}}{\pi^2} K(\Point{j'}, \Point{j})K(\Point{j}, \Point{i})\Dsurf{j'}\Dsurf{j}.
\end{equation}
and for the light rays reflected three times, by following the same reasoning line, we have:

\begin{equation} \label{eq:E3Int}
\begin{aligned}
\Illum{3}(\Point{i}) = \int_{\Surf{}}\int_{\Surf{}}\int_{\Surf{}} \Reflect{j} \Reflect{j'} \Reflect{j''} \frac{\Illum{0}}{\pi^3}  K(\Point{j''}, \Point{j'})K(\Point{j'}, \Point{j})   \\ K(\Point{j}, \Point{i})  \Dsurf{j''}\Dsurf{j'}\Dsurf{j}.
\end{aligned}
\end{equation}
and so on.

The above equations, containing integrals, cannot be analytically computed in the general case. However, by sampling the surface into a finite number of facets as proposed by Nayar \textit{et al.} \cite{Nayar91}, a discrete version of the model can be obtained, providing a simpler mathematical formalism. A surface can be sampled into $m$ small facets, each facet is assumed uniformly illuminated, flat and uniform in reflectance.
Let us denote as $K_{ij}$ the geometrical kernel defined by Equation (\ref{eq:kernel}) where $\Point{i}$ and $\Point{j}$ are the centers of facets $i$ and $j$ of respective areas $S_i$ and $S_j$. Then, according to Equations (\ref{eq:E1Int}), (\ref{eq:E2Int}) and (\ref{eq:E3Int}), the irradiance in a facet centered on $\Point{i}$, after first, second and third bounces can be re-written as:

\begin{equation} \label{eq:Esum1}
\Illum{1}(\Point{i}) = \sum_{j=1}^{m} \Reflect{j} \frac{\Illum{0}}{\pi} K_{ji} S_j,
\end{equation}

\begin{equation} \label{eq:Esum2}
\Illum{2}(\Point{i}) = \sum_{j=1}^{m}\sum_{j'=1}^{m} \Reflect{j} \Reflect{j'} \frac{\Illum{0}}{\pi^2}, K_{j'j} K_{ji} S_{j'} S_j,
\end{equation}

\begin{equation} \label{eq:Esum3}
\Illum{3}(\Point{i}) = \sum_{j=1}^{m} \sum_{j'=1}^{m}\sum_{j''=1}^{m} \Reflect{j} \Reflect{j'} \Reflect{j''} \frac{\Illum{0}}{\pi^3} K_{j''j'} K_{j'j} K_{ji}  S_{j''} S_{j'} S_{j}.
\end{equation}

These equations can be turned into a matrix formalism, allowing to consider all facets simultaneously. Let us define a symmetric matrix containing all the geometrical kernel values associated with all the pairs of facets on the surface: 

\begin{equation} \label{eq:kernel_matrix}
\mathbf{K} = \begin{bmatrix}
0 & K_{12} &. & . & K_{1m}\\
K_{21} & 0 & . & . & K_{2m} \\
. &. & 0 &. &. \\
K_{m1} & .& . & . & 0
\end{bmatrix}\
\end{equation}

Note that the zeros on the diagonal of $\mathbf{K}$ are due to the fact that rays cannot transit to a facet from itself. Let us also define a diagonal matrix $\mathbf{R}$ containing the spectral reflectances of the different facets for a single wavelength:

\begin{equation}  \label{eq:reflect_matrix}
\mathbf{R} = \begin{bmatrix}
\Reflect{1} &0 &...... & 0\\
0 & \Reflect{2} & .... & 0 \\
. & .& . & .... \\
. & .& . & ... \\
0 & 0& .. & \Reflect{m} \\
\end{bmatrix}
\end{equation}

In order to simplify the final equation, both the factor $1/\pi$ and the corresponding facet area can be included into the entries of matrix $\mathbf{K}$. Thus, the matrix $\mathbf{K}$ can be redefined as follows:

\begin{equation} \label{eq:kernel_matrix2}
\mathbf{K} = \frac{1}{\pi} \begin{bmatrix}
0 & K_{12} S_{2} &. & . & K_{1m} S_{m}\\
K_{21} S_{1} & 0 & . & . & K_{2m} S_{m} \\
. &. & 0 &. &. \\
K_{m1} S_{1} & .& . & . & 0
\end{bmatrix}\
\end{equation}

This matrix is symmetric only when all the facets are of equal size. 

Based on these definitions, Equations (\ref{eq:Esum1}), (\ref{eq:Esum2}) and (\ref{eq:Esum3}) can be written in matrix form, respectively:

\begin{equation} 
\mathbf{E_1} =   \mathbf{K}  \mathbf{R}  \mathbf{\Illum{0}},
\end{equation} 

\begin{equation} 
\mathbf{E_2} =   (\mathbf{K}  \mathbf{R})^2  \mathbf{\Illum{0}},
\end{equation}

\begin{equation} 
\mathbf{E_3} =   (\mathbf{K}  \mathbf{R})^3  \mathbf{\Illum{0}},
\end{equation}
where  $\mathbf{E_0}$, $\mathbf{E_1}$, $\mathbf{E_2}$ and $\mathbf{E_3}$ are vectors of size $m$ containing the irradiances on each of the $m$ facets, due to light having undergone $0$, $1$, $2$ or $3$ bounces respectively.  

 Then, the irradiane vector related to the irradiance values on each of the $m$ facets, after $n$ bounces of light, can be written as:

\begin{equation}  \label{eq:sumIllum}
\mathbf{E} =  \sum_{b=0}^{n}  ( \mathbf{K}  \mathbf{R})^b  \mathbf{\Illum{0}}  =  \sum_{b=0}^{n} \mathbf{E_b}.
\end{equation} 

This sum corresponds to a geometric series, which, when $n$ tends to infinity, converges to:

\begin{equation} \label{eq:illum}
\mathbf{E}  = ( \mathbf{I} -  \mathbf{K}  \mathbf{R})^{-1}   \mathbf{\Illum{0}}.
\end{equation}

The convergence is guaranteed for non fluorescent materials as shown in \ref{s:append}. 

Equation (\ref{eq:illum}) is a general expression of irradiance after infinite bounces of light for a Lambertian surface. It is a function of wavelength, as both the reflectance of the surface and the power distribution of the lighting are spectral functions. Likewise, the radiance vector containing  radiances reflected from the $m$ facets towards the camera \footnote{For Lambertian surfaces, the radiance reflected from a facet toward the camera is the same as the radiance in any other direction} is given by:

\begin{equation} \label{eq:rad}
\begin{aligned}
\mathbf{L} &= \frac{1}{\pi}  \mathbf{R} ( \mathbf{I} -   \mathbf{K}  \mathbf{R})^{-1}   \mathbf{\Illum{0}}  \\
&= \frac{1}{\pi} (  \mathbf{R}^{-1} -   \mathbf{K})^{-1}   \mathbf{\Illum{0}}.
\end{aligned}
\end{equation}

\section{Spectral interreflection model} 

In a vision system, radiance is captured by camera sensors, and converted to image intensity values, $\rho^k$ , according to the spectral response $C^k(\lambda)$ of each camera sensor, $k$, as follows:

\begin{equation} \label{eq:respInt}
\rh{k} = \int_{\lambda}  C^k(\lambda) L(\lambda)d\lambda.
\end{equation}

By sampling the light spectrum into $q$ wavebands, Equation (\ref{eq:respInt}) can be written in a matrix form: 

\begin{equation} \label{eq:rhoMat}
\boldsymbol{\rho} =    \mathbf{C}  \mathbf{L}_ {\lambda}, 
\end{equation}
where $\boldsymbol{\rho}$ is a column vector of size $s$, containing the sensor values of a pixel for each of the $s$ channels, $\mathbf{L}_ {\lambda}$ is a column vector of size $q$ containing the radiance values of the $q$ different wavebands, and $ \mathbf{C}$ is the camera response matrix of size $s \times q$. 

Notice that Equation (\ref{eq:rhoMat}) is defined for one pixel and all wavelengths, whereas Equation (\ref{eq:rad}) is defined for one wavelength and all pixels. Thus, in order to combine the two equations and to take into consideration all pixels and all wavebands at the same time, extending the matrices in both equations is needed. The vector $\boldsymbol{\rho}$ is extended to vector $\boldsymbol{\rho}_{ext}$ whose length is $ms$, where $m$ is the number of pixels in the image, and whose form is:

\begin{equation}
\boldsymbol{\rho}_{ext} = \begin{bmatrix}
\rh{1}{1} \quad . .  \quad \rh{1}{m} \quad \rh{2}{1} \quad ..  \quad \rh{2}{m} \quad  ... \quad \rh{s}{1} \quad .. \quad \rh{s}{m}
\end{bmatrix}^T,
\end{equation}
where superscript $T$ denotes the matrix transpose operator, and $\rh{
j}{i}$ is the response of channel $j$ at the pixel $i$ .

The irradiance vector $\mathbf{\Illum{0}}$ is extended in a similar way. The extended form, $\mathbf{\Illum{0}}_{ext}$,  of length $mq$, is obtained by concatenating the $\mathbf{\Illum{0}}$ vectors for each wavelength the one after the other.

The reflectance matrix $\mathbf{R}$ is extended to the square diagonal matrix $\mathbf{R}_{ext}$ of size $mq \times mq$. As $\mathbf{R}$ is defined for one wavelength, we can name it $\mathbf{R}_\lambda$, then the extended form is a concatenation of all diagonal matrices, $\mathbf{R}_\lambda$, on the diagonal of the new matrix: 

\begin{equation}
\mathbf{R}_{ext} = \begin{bmatrix}
\mathbf{R}_{\lambda_1} &0 &...... & 0\\
. & \mathbf{R}_{\lambda_2}& ..... & . \\
. &  .... & . & 0 \\
0 & .& ...... & \mathbf{R}_{\lambda_q}\\
\end{bmatrix}.
\end{equation}

The geometrical kernel matrix $\mathbf{K}$ is extended to a $mq \times mq$ block matrix whose blocks on the diagonal are $\mathbf{K}$ and all others are zero block matrices:

\begin{equation}  \label{eq:K_ext}
\mathbf{K}_{ext} = \begin{bmatrix}
\mathbf{K} &0 &...... & . & . & . &.& 0\\
. & \mathbf{K}& . & .... & . & . &.& . \\
0 & 0& .. & .  & . & . &...... & \mathbf{K}\\
\end{bmatrix}.
\end{equation}

Finally, the camera response matrix is extended to the size $ms \times mq$ to take into consideration all the pixels. In order to explain the form of matrix $\mathbf{C}_{ext}$, let us introduce the matrix $\mathbf{C^i}_\lambda$ for a channel, $i$, of size $m \times m$:

\begin{equation}
\mathbf{C^i}_\lambda = \begin{bmatrix}
c^i_{\lambda} &0 &...... & . & . & . &.& 0\\
0 & c^i_{\lambda}& . & .... & . & . &.& . \\
0 & 0& .. & .  & . & . &...... & c^i_{\lambda}\\
\end{bmatrix}.
\end{equation}

The extended matrix form of $\mathbf{C}$ can then be written as:

\begin{equation}
\mathbf{C}_{ext} = \begin{bmatrix}
\mathbf{C^1}_{\lambda_1} & \mathbf{C^1}_{\lambda_2} &...... & . & . & . &.& \mathbf{C^1}_{\lambda_q}\\
\mathbf{C^2}_{\lambda_1} & \mathbf{C^2}_{\lambda_2} &...... & . & . & . &.& \mathbf{C^2}_{\lambda_q} \\
. & .& . & .  & . & . &. & .\\
\mathbf{C^s}_{\lambda_1} & \mathbf{C^s}_{\lambda_2} &...... & . & . & . &.& \mathbf{C^s}_{\lambda_q} \\
\end{bmatrix}.
\end{equation}

After the introduction of these new matrices, a generalized spectral form of Equation (\ref{eq:rad}) can be written as:

\begin{equation}   \label{eq:spectral}
\boldsymbol{\rho}_{ext} =  \frac{1}{\pi} \mathbf{C}_{ext}  ( \mathbf{R}^{-1}_{ext} -  \mathbf{K}_{ext})^{-1}  \mathbf{\Illum{0 ext}}.
\end{equation}

\section{Spectral reflectance estimation} \label{spectEstimation}

Now that we have related RGB values of pixels in an image of a concave surface to the spectral reflectance and the geometry of the surface, the illuminant SPD, and the sensor response functions, we propose to use the interreflection model in an inverse approach in order to estimate the spectral reflectance of a surface, i.e. matrix $\mathbf{R}_{ext}$, assuming that all other matrices and vectors in Equation (\ref{eq:spectral}) are known. 

 The problem can be formulated as a constrained minimization one, and the objective function can be written as:

\begin{equation} \label{eq:minimGeneral}
\begin{split}
\mathbf{\widehat{R}}_{ext} = \argmin_{\mathbf{R}_{ext}}(| \boldsymbol{\rho}_{img} -  (\frac{1}{\pi} \mathbf{C}_{ext} ( \mathbf{R}^{-1}_{ext} -  \mathbf{K}_{ext})^{-1}  \mathbf{\Illum{0 ext}}) |^2 + \\ 
\alpha | \frac{\delta^2 r(\lambda)}{\delta \lambda^2} |^2 ), 
\text{ subject to } \forall r, 0 < r  \leq  1, 
\end{split}
\end{equation}
where $\delta^2 r(\lambda)/\delta \lambda^2$ is a smoothness constraint controlled by the smoothness parameter $\alpha$. This constraint helps to remove numerical noise and to obtain a spectrum that has a smoother shape. This choice is widely adopted in optimization, and is motivated by the observation that natural reflectances tend to be smooth rather \cite{park07}. 
 
\begin{figure}[ht]
	\centering
	\includegraphics[width=0.25\textwidth]{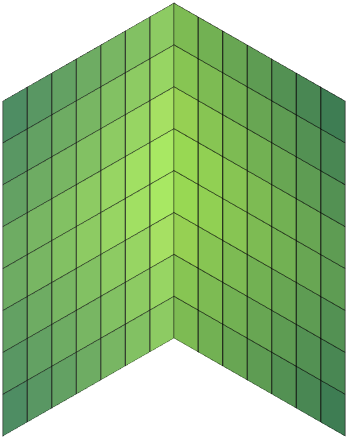}
	\caption{Interreflection effect: simulated RGB colors of a folded green patch from the Macbeth Color Checker, discretized into 8x14 facets.  }
	\label{fig:interEffect}
\end{figure}

As seen before, existing approaches in the state of the art need to use multiple light sources, and to reduce the dimensionality of the problem by using basic functions in order to estimate spectral reflectance from RGB values. In our case, the use of a single image of the surface under one illuminant is made possible thanks to interreflections. The effects of these interreflections are more or less pronounced in the different areas of the surface, as illustrated in Figure \ref{fig:interEffect}. Thus, the camera output of a single folded surface contains many different colors which provides a number of equations sufficient to solve for the unknowns even without the use of basis functions.

Thus, one of the main differences between our approach and the state of the art ones is that interreflections create a collection of different color values that are related in a non linear way to the spectral reflectance of the surface and can be used to solve an equation of a high dimension. Our model needs to analyze the whole area of the surface, whereas in approaches relying on flat surfaces, analyzing the color of one pixel (or a mean value taken over a small area) suffices.

Since this problem is constrained and non-linear, it cannot be defined mathematically as an inverse problem \cite{heikkinen07}. The number of unknowns to estimate is related to the number of different spectral reflectances contained in the surface, if the latter is not uniform, and to the number of wavebands considered in the visible spectrum. Solving this problem in its general form is not an easy task for multiple reasons. Firstly, estimating the number of unknowns related to this equation is not straightforward given the non-linear nature of the model. Secondly, if different facets in the surface have different spectral reflectances, then there is no unique solution for Equation \ref{eq:minimGeneral}. For these reasons, we propose to focus on a special case where the surface has a homogeneous spectral reflectance over its whole area. 

\subsection{Uniformly colored surface}

By focusing on the special case of a uniformly colored concave surface, a simplified model can be obtained giving a new version of the objective function. 

First, equation (\ref{eq:rad}) can be written as:

\begin{equation}  
\mathbf{L} = \frac{1}{\pi} (  \frac{1}{r}\mathbf{I} -   \mathbf{K})^{-1}   \mathbf{\Illum{0}}, 
\end{equation}
where $\mathbf{I}$ is the identity matrix of size $m \times m$ and $r$ the reflectance of the surface at the considered wavelength, assumed to be non-zero.

Assuming a discretization into equal size facets, $\mathbf{K}$ is then symmetric and the formal expression can be rewritten based on \textit{Eigendecomposition} as:

\begin{equation}   
\mathbf{L} = \frac{1}{\pi} (  \frac{1}{r}\mathbf{I} - \mathbf{Q} \mathbf{G} \mathbf{Q}^{-1}   )^{-1}   \mathbf{\Illum{0}}, 
\end{equation}
where $\mathbf{Q}$ contains the eigenvectors of $\mathbf{K}$, and $\mathbf{G}$ is a diagonal matrix containing the eigenvalues of $\mathbf{K}$.

By developing this equation, we can write:

\begin{equation}   
\mathbf{L} = \frac{1}{\pi} \mathbf{Q}  (\frac{1}{r}\mathbf{I} - \mathbf{G})^{-1} \mathbf{Q}^{-1}  \mathbf{\Illum{0}}
\end{equation}	

As explained in the previous section, the extended version of matrix $\mathbf{K}$, given by Equation (\ref{eq:K_ext}), is organized into blocks on the diagonal. Then, the inverse of matrix $ ( \mathbf{R}^{-1}_{ext} - \mathbf{K}_{ext})$, denoted as $\mathbf{M}$, can be written as:

 \begin{equation}  
 \mathbf{M} = 
 \begin{bmatrix}
 \mathbf{Q}  (\frac{1}{r_1}\mathbf{I} - \mathbf{G})^{-1} \mathbf{Q}^{-1}  &..& 0\\
 . &  .. &. \\
 0 & .. & \mathbf{Q}  (\frac{1}{r_m}\mathbf{I} - \mathbf{G})^{-1} \mathbf{Q}^{-1} \\
 \end{bmatrix},
 \end{equation}
 which can be further simplified as:

 \begin{equation}  
\mathbf{M} = \mathbf{Q}_{ext}
\begin{bmatrix}
  (\frac{1}{r_1}\mathbf{I} - \mathbf{G})^{-1}   &..& 0\\
. &  .. &. \\
0 & .. &  (\frac{1}{r_m}\mathbf{I} - \mathbf{G})^{-1}  \\
\end{bmatrix}  \mathbf{Q}^{-1}_{ext},
\end{equation}
where

\begin{equation}
\mathbf{Q}_{ext} = \begin{bmatrix}
\mathbf{Q} &0 &...... & 0\\
. & \mathbf{Q}& ..... & . \\
. &  .... & . & 0 \\
0 & .& ...... & \mathbf{Q}\\
\end{bmatrix},
\end{equation}
and,

\begin{equation}
\mathbf{Q}^{-1}_{ext} = \begin{bmatrix}
\mathbf{Q^{-1}} &0 &...... & 0\\
. & \mathbf{Q^{-1}}& ..... & . \\
. &  .... & . & 0 \\
0 & .& ...... & \mathbf{Q^{-1}}\\
\end{bmatrix}.
\end{equation}

Equation (\ref{eq:spectral}) can be rewritten as:

\begin{equation} \label{eq:modelSimp}
\boldsymbol{\rho}_{ext} =  \frac{1}{\pi} \mathbf{A} \mathbf{B},
\end{equation}
 where

\begin{equation} 
\mathbf{A} =   \mathbf{C}_{ext} \mathbf{Q}_{ext} \diag(\mathbf{Q}^{-1}_{ext}  \mathbf{\Illum{0}}),
\end{equation}
 and

\begin{equation} 
\mathbf{B} =   \begin{bmatrix}
\vect(\frac{1}{r_1}\mathbf{I} - \mathbf{G})^{-1}) \\
.   \\
.   \\
\vect(\frac{1}{r_l}\mathbf{I} - \mathbf{G})^{-1})  \\
\end{bmatrix},
\end{equation}
where $\vect$ is the vectorization operator. 

Note here that the calculation of the values in vector $\mathbf{B}$ is straightforward: as both $\mathbf{G}$ and $\mathbf{I}$  are diagonal matrices, the inverse of the matrix $(\frac{1}{r_l}\mathbf{I} - \mathbf{G})$ can be obtained by a simple division.
  
Finally, since the surface is uniform, its spectral reflectance can be fully characterized by one vector $\mathbf{R}_{\lambda}$ of size $q$. The minimization problem formulated in Equation (\ref{eq:minimGeneral}) becomes:

\begin{equation}  \label{eq:optimSimp}
\begin{split}
\mathbf{\widehat{R}}_{\lambda} = \argmin_{\mathbf{R}_{\lambda}}(| \boldsymbol{\rho}_{img} -  \frac{1}{\pi} \mathbf{A} \mathbf{B} |^2 + \alpha | \frac{\delta^2 r(\lambda)}{\delta \lambda^2}), \\ 
\text{ subject to } \forall r, 0 < r  \leq  1. 
\end{split}
\end{equation}

\subsection{Implementation}

The form of the minimization problem described by Equation (\ref{eq:optimSimp}) is very handy: only one matrix inverse ($\mathbf{Q^{-1}}$) is needed and can be calculated before starting the minimization process. Our problem is still constrained and non-linear, but the number of variables to be estimated depends only on the number of wavebands considered. We can assume that this problem is convex, and that a local minimum optimization algorithm is able to give a good solution. A more detailed study of the nature of this problem will be explained in Section \ref{discussion}. 

The minimization method we chose is the interior point algorithm \cite{kojima89,byrd99}. This method is implemented under the function \textit{fmincon} function in \textbf{Matlab(R)} software by Mathworks. The minimization is done iteratively. 

\section{Experiments and results} \label{results}

In order to verify the validity of the proposed model and its capacity to provide accurate spectral reflectances of Lambertian surfaces, we carried out various tests, first based on simulated images in order to estimate the optimal performance of the method, then based on real images in order to see the influence of the noise in the image and other geometrical imperfections.

\subsection{Case study}
For our tests on both synthetic and real camera data, we considered the following case study: a scene consisting of a Lambertian surface of uniform spectral reflectance, folded into a V-shaped surface of two similar planar square panels $\Surf{1}$ and $\Surf{2}$ with an angle of $45^\circ$ between them. Each panel is discretized into $100$ equally-sized facets. The geometrical kernel is then calculated using Monte Carlo estimation. Although, the results presented here are for a specific configuration, the effect of the angle and the discretization size will be studied later in Section \ref{discussion}. The V-cavities are illuminated frontally by collimated lighting parallel to the bisecting plane of the two panels, thus every facet is assumed to receive the same amount of lighting.


\subsection{Evaluation}

The performance of the spectral reflectance estimation is evaluated using the root mean square error (RMSE) calculated between the estimated spectral reflectance $\widehat{\mathbf{R}}$ and the ground truth one $\mathbf{R}$ \footnote{The ground truth spectral reflectance is the one used in the simulation in case of experiments on synthetic data. For real data, ground truth spectral reflectances are measured using a spectrometer. }:

\begin{equation}  \label{eq:rmse}
RMSE(\mathbf{R},\widehat{\mathbf{R}}) = \sqrt{|\mathbf{R} - \widehat{\mathbf{R}}|^2 /q  },
\end{equation}
where $q$ is the number of wavebands.

The performance is also evaluated in terms of color distance by using  CIEDE00 distance \cite{luo01},  computed between the ground truth spectral reflectance, $\mathbf{R}$, and the estimated one, $\widehat{\mathbf{R}}$, for a $10^\circ$ standard observer and a CIE D65 illuminant.

For real camera output data, we added the Pearson distance (PD) to the used metrics:

\begin{equation}  \label{eq:rmse}
PD(\mathbf{R},\widehat{\mathbf{R}}) = 1 - \frac{\mathbf{R}^T \widehat{\mathbf{R}}}{|\mathbf{R}| |\widehat{\mathbf{R}}|}.
\end{equation} 

This distance is independent of the magnitude, thus it helps in giving better evaluation based on the shape of the estimated spectral reflectance especially in case of non calibrated settings.

\subsection{Synthetic data}

We started by testing our method on synthetic data. For this sake, scenes with interreflections were simulated supposing a collimated light source with SPD corresponding to CIE D65 illuminant, a patch with a spectral reflectance of one of the Mackbeth ColorChecker patches, and a given geometry of V-shape. A SD-10 camera is considered whose response curves are similar to the ones used in \cite{khan13}. Under these configurations, the RGB values for each of the patches of Macbeth ColorCheckers were obtained using our model [see Equation (\ref{eq:modelSimp})].  Afterwards, these RGB values are used to estimate back the surface spectral reflectance.

Our results are compared with the results of two other comparable approaches \cite{park07} and \cite{khan13} relying on three different illuminants. We implemented these approaches and used the first 8 components of the Parkkinen basis as suggested in \cite{park07}. For our approach, we argue that there is no more need for reducing the dimensionality of the problem, thus we estimate the spectral reflectance from $400$ nm to $700$ nm in steps of $5$ nm, giving a total of $61$ values to be estimated. The size of the planar surfaces is set to $2 \times 2$ cm and the smoothness parameter, $\alpha$, is chosen to be $2.5$. 

The resulted RMSE and CIEDE00 values shown in Table \ref{table:res} indicate that spectral reflectance is as accurately predicted by using the interreflection approach with one light as by using approaches based on several lightings, and it can be even more accurate. It is worth mentioning that using three different lights means performing the acquisition three times and using three lamps. 

\begin{table*}[htbp]
	\begin{center}
	\caption{Reflectance estimation accuracy for synthetic data}
	\label{table:res}
  \begin{tabular}{ l c c c c c  }
	\hline
	\multirow{2}{*}{Method} & \multirow{2}{*}{No. Illuminants} &  \multicolumn{2}{ c }{ ColorChecker24} &  \multicolumn{2}{ c}{ ColorChecker240} \\ 
    &    & RMSE  & CIEDE00 & RMSE & CIEDE00 \\ 
 \hline
Park \cite{park07} &3& 0.024 & 0.31 & 0.018 & 0.37 \\ 
 Khan \cite{khan13} &3 & 0.018 &  0.31 & \textbf{0.011 }& 0.30 \\ 
Our Method  & 1 & \textbf{ 0.017}  & \textbf{0.29} & 0.017 & \textbf{0.26} \\ 
\hline
\end{tabular}
\end{center}
\end{table*}

\subsection{Real data}

In this section, the results of our method on real camera output are presented. Camera outputs are taken under daylight and a standard illuminant SPD is used to simulate this light in the equations. 

\begin{figure}[ht]
	\centering
	\includegraphics[width=0.3\textwidth]{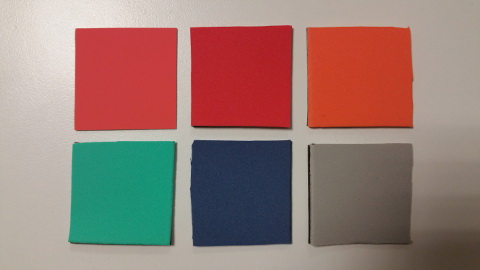}	
	\caption{The set of used samples: the top-left sample is a Munsell paper (5R 6/12 Mat). The other samples are pieces of fabrics.  }
	\label{fig:figSamples}
\end{figure}

To perform these experiments, we used a platform enabling us to fix two pieces of the same sample in a planar configuration and to set approximately an angle of $45^\circ$  between them. We used six uniformly colored samples, one of them is a Red Munsell sample (5R 6/12), and the other five are textile samples shown in Figure \ref{fig:figSamples}. For each sample, we have a set of two patches of size $4 \times 4$ cm. The ground truth spectral reflectances of these samples were measured using the Minolta X-Rite Color i7. For the acquisition, we used a Canon EOS 1000D camera with known spectral response functions to capture images in RAW format. The photos were taken under a sunny daylight in the early afternoon. Having in mind to propose a practical setup, we did not measure the spectral power distribution of this illuminant, instead we used the spectral power distribution of CIE D50 standard illuminant in the equations to represent the direct sunlight \cite{ohta06}. The intensity of the illuminant was not measured as well and the camera is used with various settings and focal lengths without any extra calibration process.    

Under the same settings, we also captured an image of all the available samples but this time without interreflection (flat samples). 

\begin{figure}[ht]
	\centering
	\includegraphics[width=0.3\textwidth]{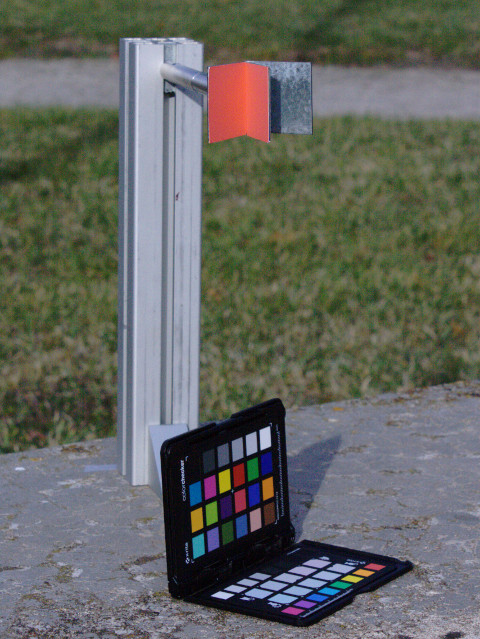}	
	\caption{Example of acquisition: The Munsell V-cavity on the top and the X-Rite Color Checker on the bottom under direct sun light.}
	\label{fig:figSamMuns}
\end{figure}

Before acquisition, a X-Rite Color Checker was added to each scene. Later, it was used to pre-calibrate the images when the tests with pre-calibration are performed. An example of a camera output used in the experiments is shown in Figure (\ref{fig:figSamMuns}).

RGB values of the image are extracted from an area selected by the user and corresponding to a whole area of one panel. A  discretization size of $4 \times 4$ mm is used to divide the selected area into $100$ facets, each one attached to an RGB value corresponding to the mean RGB value over its area. Based on this discretization size, the geometrical kernel can be built using Monte Carlo estimation method according to the panels size and the angle between them. 

  We also tested the methods described in \cite{park07} and \cite{khan13} on images of similar flat samples while using only CIE D50 light and $3$ basis functions. The RGB value given to these approaches is the mean value over an area selected by the user and corresponding to the surface of the patch.
  
  The used camera is not calibrated to the specific acquisition settings, and the intensity of light source is not measured. For these reasons, we separate the experiments into two categories:
  
  \begin{itemize}
  	\item With pre-calibration: we used the $24$ patches of the Color Checker chart to learn second degree polynomial curves relating the expected red, green and blue values with the simulated red, green  and blue ones respectively. Then, the learned transformation is applied on the image values used in the estimation. 
  	\item Without pre-calibration: in this case, no pre-calibration is used to estimate the quantity of light entering the camera lens. Instead, we propose to normalize image values and simulated values on their respective sums in order to assume that they have similar amounts of energy. In order to apply these normalization steps, we reimplemented the approaches \cite{park07} and \cite{khan13} using the same optimization algorithm as ours.
  
  \end{itemize}
  
   Tables \ref{table:resCal} and \ref{table:resNonCal} show the results of our approach compared to \cite{park07} and \cite{khan13}. Table \ref{table:resCal} shows the mean values over all the patches of RMSE, DE00 and PD in case of pre-calibration. Table \ref{table:resNonCal} shows the mean values over all the patches of RMSE, DE00 and PD when no pre-calibration is used. The results show that our approach performs better than the approaches  \cite{park07} and \cite{khan13} in both non-calibrated and pre-calibrated cases. Moreover, our results without pre-calibration seems comparable to the results of the two other approaches after pre-calibration. This shows that one can take a camera whose spectral response functions were measured previously in a lab, and use it in a complete non-calibrated settings, and use a standard SPD of light instead of measuring it and still get good spectral estimations with out model.  
   	  
  Table \ref{table:resReal} shows the results of our approach compared to \cite{park07} and \cite{khan13} for each of the used samples. On the one hand, the results of our approach are obtained without any pre-calibration. On the other hand, when applying the approaches \cite{park07} and \cite{khan13}, the pre-calibration step using the Color Checker chart was applied. By observing Table \ref{table:resReal}, one can see that in most cases our approach performed better even though no pre-calibration is performed in our case. An exception can be seen for two particular patches, the gray and the blue ones, where the surface is weakly reflective. Actually, interreflections help in spectral reflectance estimation only in case of surfaces with high reflectance in some wavebands. This is due to the fact that interreflections are very weak when the surface reflectance is low \cite{deeb17}. 
  
\begin{figure}[!h]
	\centering
	\begin{subfigure}[b]{0.4\textwidth}
		\includegraphics[width=\textwidth]{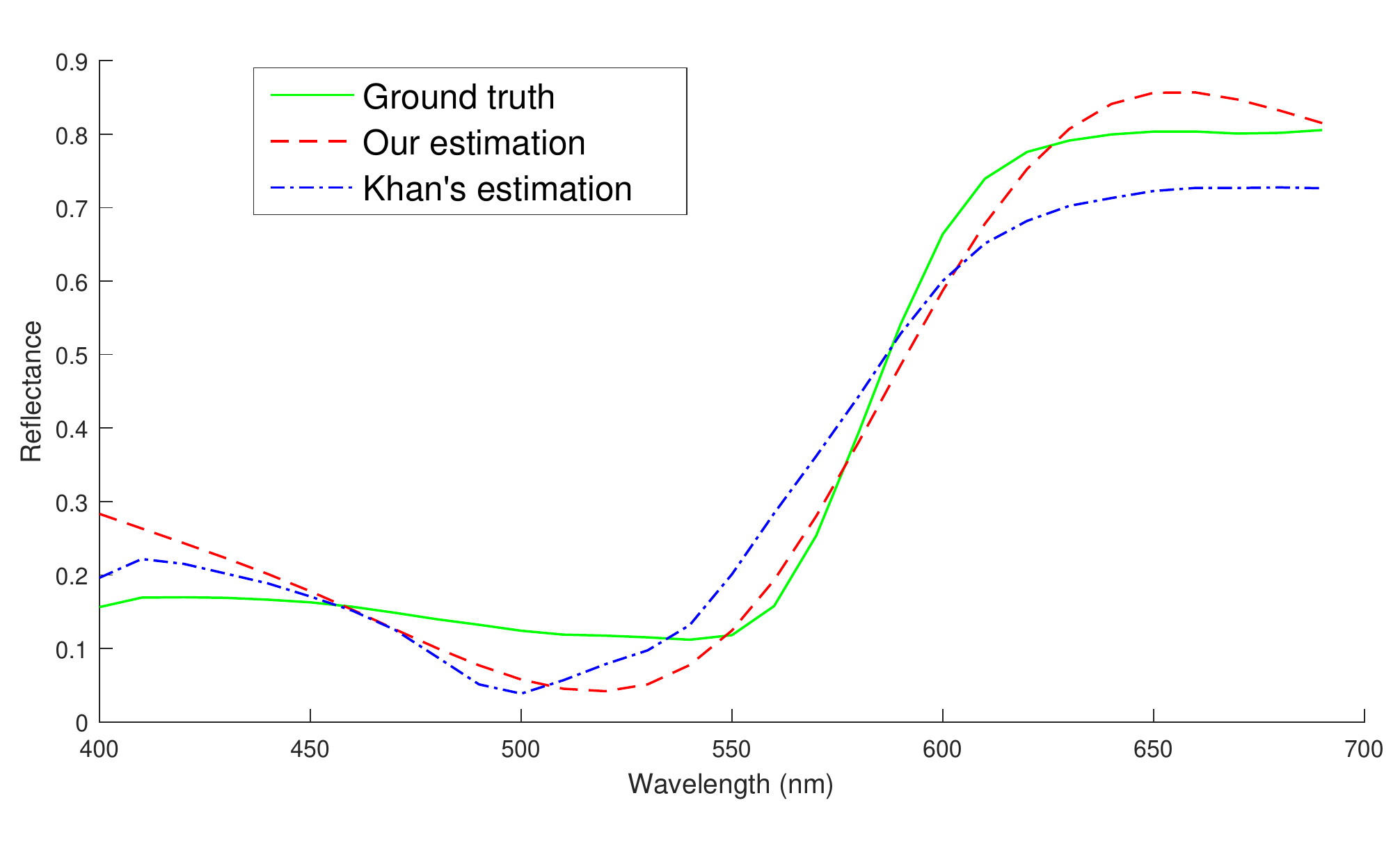}
		\caption{A comparison of estimation for Munsell red patch }
		\label{fig:RMuns}
	\end{subfigure}
	
	\begin{subfigure}[b]{0.4\textwidth}
		\includegraphics[width=\textwidth]{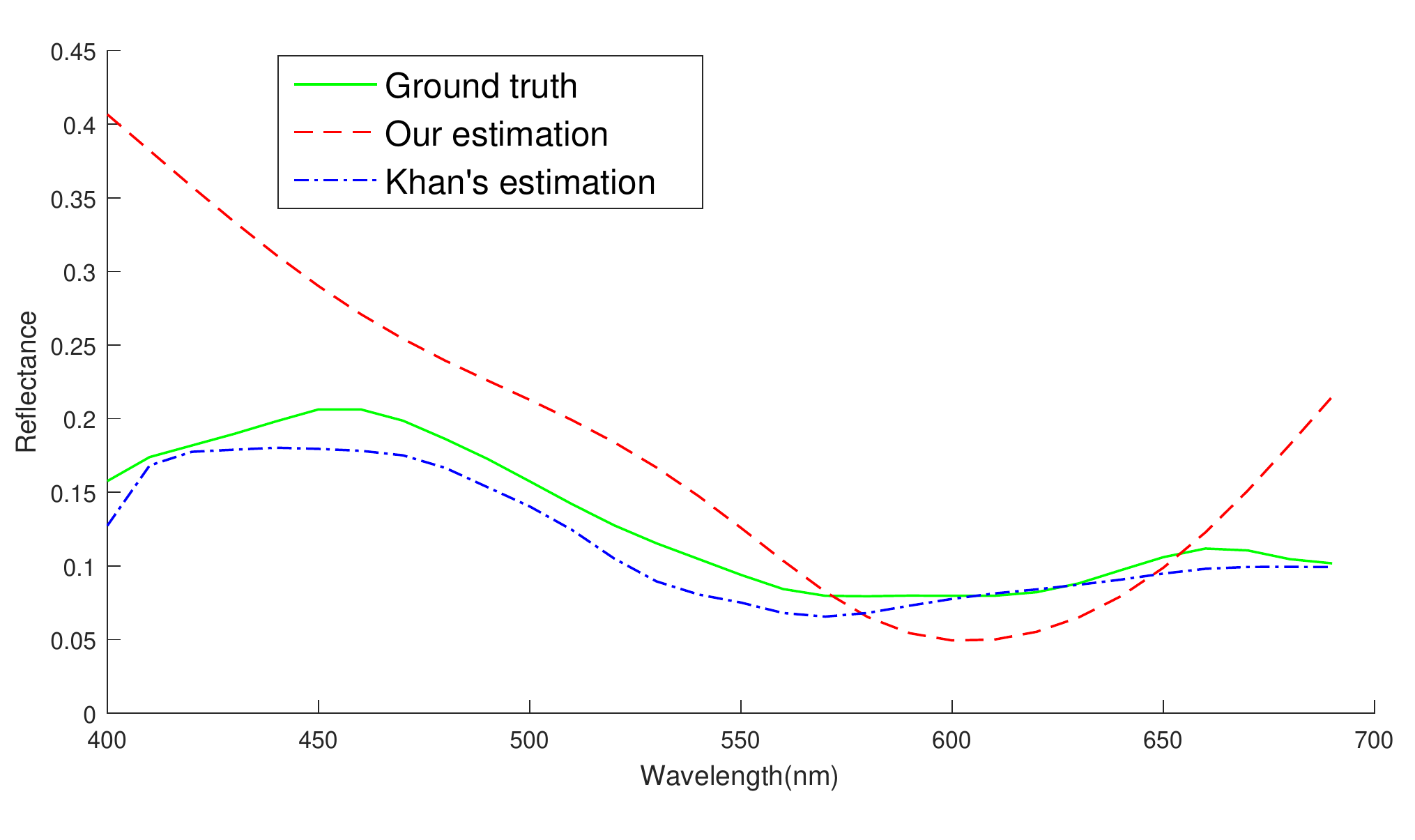}
		\caption{A comparison of estimation for blue patch}
		\label{fig:RBlue}
	\end{subfigure}
	
	\caption{Comparison between the estimated spectral reflectences and the ones measured with a spectrometer of some surfaces: in green the ground truth, in red the estimation based on our method without pre-calibration, in blue the estimation based on Khan et al. approach \cite{khan13} after pre-calibration.}\label{fig:compR}
\end{figure}

   The estimated spectral reflectances compared to the ground truth ones are shown in Figure (\ref{fig:compR}). In (\ref{fig:RMuns}), the graph shows that our approach without pre-calibration outperformed the approaches \cite{park07} and \cite{khan13} after pre-calibration for the case of Munsell patch. However, in the case of low reflectance surfaces, for example the blue one, our approach failed to accurately estimate its spectral reflectance, see Figure (\ref{fig:RBlue}).
   
   But why interreflections help in case of non-calibrated settings? To answer this question, consider two different cases. In the first case, let us consider that the power of light used in the acquisition is multiplied by a factor $f_1$, then by a factor $f_2$. After normalization, RGB values with and without interreflections will be the same in both cases as the factors  $f_1$ and  $f_2$ are canceled by the normalization step. In the second case, consider instead two surfaces where the spectral reflectance of the first surface is $f$ times the spectral reflectance of the other one. Here, in case of flat surfaces, RGB values for both surfaces will be the same as the factor $f$ is canceled by the normalization step. However, in case of interreflections, and due to the fact that reflectance is raised to different powers with each bounce, the factor $f$ cannot be canceled by the normalization step and RGB values will differ between one surface and the other. Thus, interreflections help in distinguishing changes in spectral reflectance from changes in light power, and the variations of RGB values over the surface help in retrieving the correct spectral reflectance.

 \begin{table}[!h]
 	\begin{center}
 		\caption{Reflectance estimation accuracy on real images taken under daylight after pre-calibration}
 		\label{table:resCal}
 		\begin{tabular}{ l c c c }
 			\hline
 			Method     & RMSE  & CIEDE00 & PD  \\ 
 			\hline
 			Park \cite{park07} & 0.060 & \textbf{3.76 }& 0.009 \\ 
 			Khan \cite{khan13} & 0.059 &  3.87 & 0.009 \\ 
 			Our Method  & \textbf{ 0.046}  & 3.82 & \textbf{0.008} \\ 
 			\hline
 		\end{tabular}
 	\end{center}
 \end{table}
 
 \begin{table}[!h]
 	\begin{center}
 		\caption{Reflectance estimation accuracy on real images taken under daylight without pre-calibration}
 		\label{table:resNonCal}
 		\begin{tabular}{ l c c c }
 			\hline
 			Method     & RMSE  & CIEDE00 & PD  \\ 
 			\hline
 			Park \cite{park07} & 0.143 & 8.02 & 0.012 \\ 
 			Khan \cite{khan13} & 0.138 &  7.99 & 0.012 \\ 
 			Our Method  & \textbf{ 0.061 } & \textbf{5.62} & 0.012 \\ 
 			\hline
 		\end{tabular}
 	\end{center}
 \end{table}

\begin{table*}[!t]
	\begin{center}
		\caption{Details of reflectance estimation accuracy on real images taken under daylight}
		\label{table:resReal}
		\begin{tabular}{  l l c c c }
			\hline
			\multirow{2}{*}{Patch} & \multirow{2}{*}{Metric} & 	Park \cite{park07} &   Khan \cite{khan13} & Our Method  \\   & 
			& With pre-calibration &   With pre-calibration & Without pre-calibration  \\ 
			\hline
			\multirow{3}{*}{Munsell Red} & RMSE & 0.078 & 0.078 &\textbf{ 0.055}  \\ 
			  						 & CIEDE00 & 5.03 & 5.38 &\textbf{ 4.74} \\ 
			 						 & PD &\textbf{ 0.006} &  0.007 &  \textbf{0.006 }\\ \hline
			\multirow{3}{*}{Orange} & RMSE & 0.076 & 0.069 & \textbf{0.061}  \\ 
									 & CIEDE00 & \textbf{3.53} & 3.71 & 3.71 \\ 
			 						 & PD & \textbf{0.004} &  \textbf{0.004} &  0.006 \\ \hline
			\multirow{3}{*}{Cyan} & RMSE & 0.088 & 0.076 & \textbf{0.065}  \\ 
			 						& CIEDE00 & 3.44 & \textbf{3.40} & 5.89 \\ 
			 						& PD & 0.022 &  0.016 & \textbf{ 0.006} \\ \hline
			\multirow{3}{*}{Red} & RMSE & 0.088 & 0.10 & \textbf{0.069}  \\ 
			 						& CIEDE00 & 6.31 & 6.78 &\textbf{ 5.80} \\ 
			 						& PD & 0.019 &  0.026 &    \textbf{0.012} \\ \hline
			 \multirow{3}{*}{Blue} & RMSE & 0.009 &\textbf{ 0.008} & 0.084  \\ 
			 						& CIEDE00 & 2.88 &\textbf{ 2.77} & 8.74 \\ 
			 						& PD &\textbf{ 0.002} &  0.002 &    0.039 \\ \hline
			 \multirow{3}{*}{Gray} & RMSE & 0.022 & \textbf{0.021} & 0.035  \\ 
									& CIEDE00 & 1.35 & \textbf{1.18} & 4.82 \\ 
									& PD & \textbf{0.001} &  \textbf{0.001} &    0.004 \\  \hline			
			
		\end{tabular}
	\end{center}
\end{table*}

\section{Discussion} \label{discussion}

In this section, we propose a detailed study of the influence on the estimation accuracy of the angle between the two panels and the chosen facet size. In addition, an analysis of the mathematical nature of the problem  that we solve and the uniqueness of the solution is given. 
  
\subsection{Study of the effect of angle and facet size}

We performed an experimental study on the effect of the angle between the two panels, on the accuracy of the spectral reflectance estimation using our method. Angles of $90^\circ$, $60^\circ$, $45^\circ$ and $30^\circ$ are used. Another parameter that may have an important effect on the results is the spatial sampling of the surface (size and number of facets). Discretization into $64$, $100$ and $256$ facets on each planar surface is used. Results in term of RMSE and CIEDE00 are shown in Table \ref{table:compAng}.

\begin{table*}[ht]
	\begin{center}
		\caption{Comparison of reflectance estimation accuracy for our method when using different angles and facet sizes}
		\label{table:compAng}
		\begin{tabular}{ l c c c c c c c c c c c c }
			\hline
		Angle & \multicolumn{3}{ c }{ $30^\circ$} &  \multicolumn{3}{ c}{ $45^\circ$} & \multicolumn{3}{ c }{ $60^\circ$} &  \multicolumn{3}{ c}{ $90^\circ$} \\  
		\hline
			Number of facets & 64 & 100 & 256  & 64 & 100 & 256 & 64 & 100 & 256 & 64 & 100 & 256 \\ 
			RMSE &  0.016 & \textbf{0.0158} & 0.0160  & 0.0189 & 0.0172 &\textbf{ 0.0170} & 0.0189 & 0.0181 & 0.0182 & 0.0250 & 0.0246 &\textbf{ 0.0244} \\ 
			CIEDE00 & 0.281 & 0.287 & \textbf{0.265 } & 0.295 & \textbf{0.292} & 0.305 & 0.321 &\textbf{ 0.308} & 0.316 & 0.423 & 0.391 &\textbf{ 0.387} \\ 
			\hline
		\end{tabular}
	\end{center}
\end{table*}

\begin{figure}[!h]
	\centering
	\begin{subfigure}[b]{0.3\textwidth}
		\includegraphics[width=\textwidth]{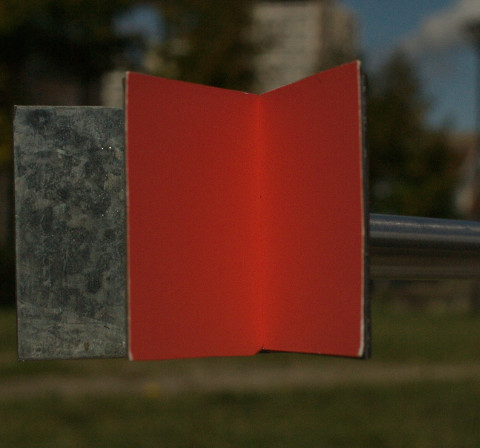}
		\caption{Two Munsell sheets with an angle of $45^\circ $ }
		\label{fig:munsell45}
	\end{subfigure}
\\
  
	\begin{subfigure}[b]{0.3\textwidth}
		\includegraphics[width=\textwidth]{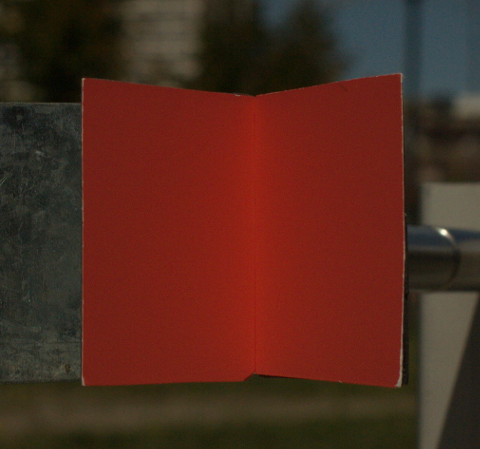}
		\caption{Two Munsell sheets with an angle of $60^\circ $ }
		\label{fig:munsell60}
	\end{subfigure}
	
	\caption{Effect of angle change on image RGB values}\label{fig:munsellAng}
\end{figure}

By observing changes of error values with angle in Table \ref{table:compAng}, it is clear that the error gets lower as the angle decreases. A smaller angle emphasizes the interreflection effects, especially near the joint border of the panels where light has more chance to bounce multiple times. In order to take advantage of this, a small discretization step is preferable to better observe the changes in RGB values per facet \ref{fig:munsellAng}. 

Nonetheless, when observing the changes of errors with facet sizes in Table \ref{table:compAng}, it is noticeable that using different facet sizes did not significantly affect the errors in general. An interpretation of this observation may be related to the dimensionality of the problem: enough information can be obtained when the two surfaces get closer to each other without the need of a very small facet size. In other words, using more facets may not necessarily help in adding significantly different RGB values to help the optimization process.  This depends on the choice of the angle and on the number of unknowns to be estimated. This question becomes more relevant in real situations, as a coarse discretization may be needed to avoid noise in the image and to reduce the computation time. If more accuracy is needed, an adaptive facet size can be adopted. However, the matrix $\mathbf{K}$ would be no more symmetric and the form proposed for the self-interreflection case in Equation \ref{eq:modelSimp} would not be valid anymore. 

Practically speaking,  angles between $45^\circ$ and $60^\circ$ are good choices. They are small enough for significant interreflections to happen, and big enough to let the surface receive sufficient amount of light. The panel should have at least the number of used facets, but it is recommended to have more in order to reduce the sensibility to image noise. The size of panels is optional: interreflections depend only on the relative size of the panels.

\subsection{Study of the uniqueness of solution}

The experiments reported in this study showed, empirically, the capacity of our interreflection-based method to provide rather good estimate of the spectral reflectance from one RGB image. However, since this method is neither a direct spectral reflectance measurement nor a direct combination of spectral measurements yielding a spectral reflectance, but the result of a mathematical optimization from RGB values, we must question the capacity of this operation to provide the expected spectral information, regardless of the experimental uncertainties. The question is not on the objective function that is optimized [see Equation \ref{eq:optimSimp}], which relies on a well-established physical ground through the interreflection equation. The question is rather on the convergence of the optimization process, even though we could verify in all our tests that the iterative optimization algorithm is not sensible to the starting values. In absence of mathematical proof of convergence, which looks to be a hard mathematical problem, we cannot exclude the possibility that the optimization provides a spectrum totally different from the expected spectral reflectance. This is why we prefer the term "estimation" rather than "prediction" or "measurement". However, physical arguments help us to claim that the results of the optimization process is most often strongly correlated with the surface's spectral reflectance, which means that the range of spectral reflectances that the method would systematically fail to retrieve with reasonable tolerance is rather limited, and excludes most real materials. 

In order to understand why the method has good chance to converge towards the expected result, let us remind how interreflections intrinsically bring spectral information into the color image, especially in the case of V-shaped surfaces where color gradients are well visible (actually, under directional illumination, color gradients are always displayed except in the very special case of integrating spheres where the geometrical extent between every point and every other point is constant, which results in a homogeneous color over the whole sphere and therefore an absence of color gradient). The color gradient comes from the fact that the radiances perceived from the different points of the surface contain different populations of photons having bounced different numbers of times in average in the concavity before reaching the sensor: near the external edges of the panels, photons have more chance to escape the concavity after one bounce (these photons form a flux proportional to the incident spectral irradiance and the reflectance of the surface), whereas near the joint-edge, they have more chance to strike the surface again after each bounce, therefore to bounce many times (the corresponding flux is proportional to the incident spectral irradiance and a combination of the reflectance, its square, its cube, and so on…). Hence, the radiance  $L(\lambda)$ viewed from a given point $P$ can be approximated as: 

\begin{equation}
	L_P(\lambda) = \sum_{j=1}^{n} q_j r^j(\lambda)
\end{equation}
where $r(\lambda)$  denotes the surface reflectance, $ q_j$  are factors related to the surface geometry, and $n$ is the highest significant number of bounces undergone by the photons which transit from point $P$ to the sensor. The number $n$ increases from the external edges where it is typically around $1-2$, to the joint edge between the panels where it can reach extremely high values. Since the spectral radiance is a non-linear function of $n$, the collection of spectral radiances viewed over the surface form a spectral space with appreciable dimensionality (which explains why there is no need with this method to use various illuminants or decrease the dimensionality of the spectral space of the considered samples). 

\begin{figure}[!h]
	\centering
	\begin{subfigure}[b]{0.4\textwidth}
		\includegraphics[width=\textwidth]{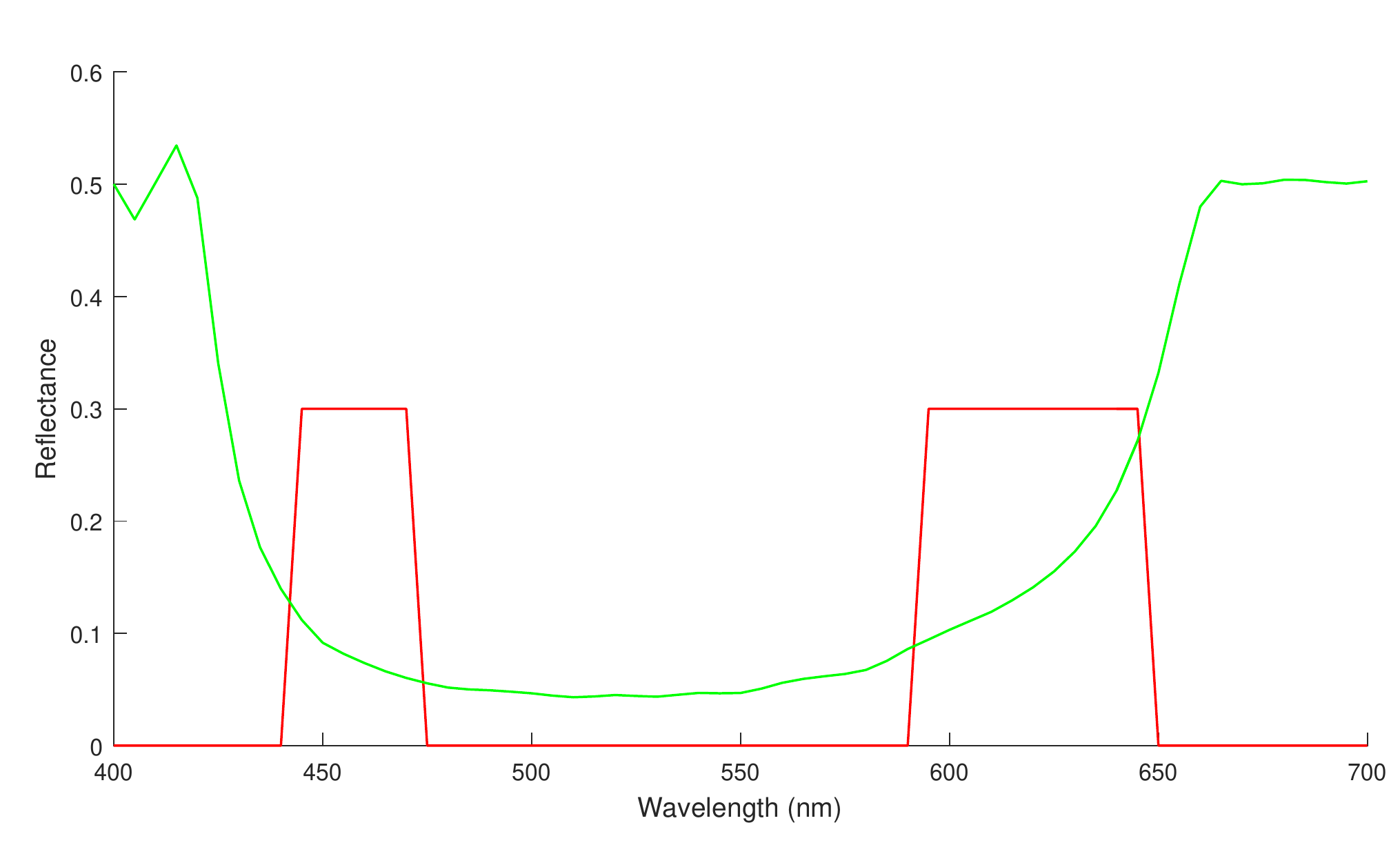}
		\caption{Spectral reflectances giving the same RGB values when the surfaces are flat}
		\label{fig:RMetar}
	\end{subfigure}
	
	\begin{subfigure}[b]{0.4\textwidth}
		\includegraphics[width=\textwidth]{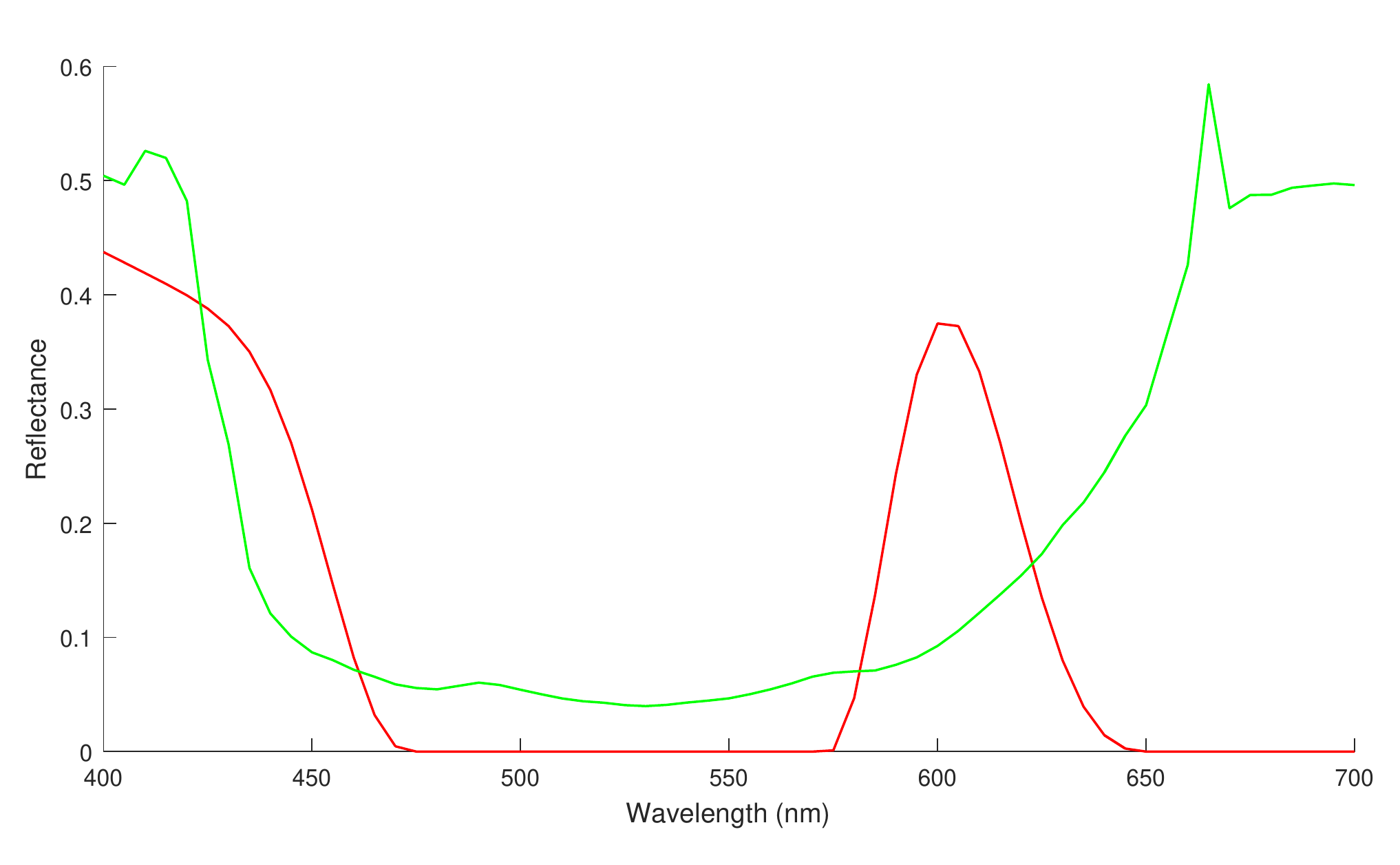}
		\caption{Spectral reflectance estimation for both spectra: in Red the estimation for the step-shaped spectrum, in green the estimation for the smooth one $\alpha$}
		\label{fig:ROptim}
	\end{subfigure}
	
	\caption{Experimental study on the uniqueness of the solution.}\label{fig:expUniq}
\end{figure}

Once captured by the sensors, the collection of spectral radiances is converted into a collection of RGB values, all correlated with the spectral radiance of the surface. If we modify the spectral reflectance of the surface, it is possible that the RGB values given by the sensors in one point are not modified (case of metamerism), but it is almost impossible that the R, G, and B values remain unmodified for all the points, even in very special cases of spectral reflectances as shown in Figure  (\ref{fig:RMetar}) through a multi-rectangular function. The two experimental spectral reflectance functions in Figure (\ref{fig:RMetar}) give exactly the same camera response with the SD-10 camera under CIE D65 illuminant. Using our model, we obtained the RGB values corresponding to each of these reflectences for a surface bent with an angle of $60\deg$. Then,  we used our objective function to find back the spectral reflectances. This is done without using the smoothness term in order to not affect the solution. Figure (\ref{fig:ROptim}) shows the estimated spectral reflectance for both these spectral reflectances. 

Therefore, the collection of RGB values displayed by an image of surface with known geometry, illuminant and sensor, seems to be a signature of the spectral reflectance of the surface: it is almost impossible to find a surface with different spectral reflectance that would display similar set of RGB colors in the same experimental configuration. This means that two metameric flat surfaces cannot be metameric anymore once bent in concave shape.

\section{Conclusion} \label{conclusion}

In this paper, we presented a study of the effect of interreflections on spectral reflectance estimation. The fact that a surface is concave leads its surface elements to reflect light towards each other so the light bounces multiple times between them. The number of bounces and their importance are defined by the geometry of the scene, and the location of each surface element in	 the object. This is the reason why different pixels of the same uniformly colored surface in an image may have different RGB values when interreflections occur. The results of our experiments showed that when using a uniformly colored surface, the whole spectral reflectance can be found with a good precision from a single RGB image of a concave surface taken under direct daylight. This performance is achieved without the need of any pre-calibration or specific acquisition settings. In addition, we presented a detailed study of the effect of the angle and the facet size on the quality of spectral reflectance estimation. Furthermore, the uniqueness of the solution has been analyzed.

The proposed solution is valid in case of collimated homogeneous light without taking into consideration the ambient light. We think that the model can be further enhanced by treating ambient light as well as direct light. Moreover, the surfaces used in the experiments were not fully Lambertian, they show some specularity at grazing incidence of light. Thus, providing a more general form of the interreflections model could be appreciable in order to enhance the accuracy of the method in real scenarios.

\section*{Appendix A: Proof of convergence of the geometric series of irradiance}   \refstepcounter{append}
\label{s:append}
The irradiance vector after $n$ bounces of light can be written as a sum of irradiance after one bounce, two bounces, three bounces,..., and $n$ bounces of light as expressed in Equation (\ref{eq:sumIllum}). This sum is a geometric series of the form $
a + ar + ar^2 + ... + ar^n$. Thus, Equation  (\ref{eq:sumIllum}) can be written as:

\begin{equation}
\mathbf{E} = (\mathbf{I}  -  (\mathbf{K}  \mathbf{R})^n) (\mathbf{I}  -  \mathbf{K}  \mathbf{R})^{-1}  \mathbf{\Illum{0}} .
\end{equation}

Then, when $n$ tends to infinity, the irradiance vector is written as expressed in Equation (\ref{eq:illum}):

\begin{equation}
\mathbf{E} =  (\mathbf{I}  -  \mathbf{K}  \mathbf{R})^{-1}  \mathbf{\Illum{0}},
\end{equation}
which can be shown by proving that $\mathbf{I}  -  (\mathbf{K}  \mathbf{R})^\infty$ tends to the identity matrix, or that $(\mathbf{K}  \mathbf{R})^\infty$ tends to the zero matrix. 

A nonnegative matrix, $\mathbf{A}$, is said to be \textit{substochastic} if it satisfies the two following conditions:
\begin{itemize}
	\item $\sum_{j} a_{ij} 	\leq 1$
	\item There is at least one element in each column that is strictly superior to zero. 
\end{itemize}

It has been proven in \cite{Hebert06} that the spectral radius of \textit{substochastic} matrices is strictly less than one and that for a \textit{substochastic} matrix $\mathbf{A}$:

\begin{equation}
\lim_{n\to\infty} \mathbf{A}^n = 0
\end{equation}

In our case, $\mathbf{R}$ is a square diagonal matrix containing the spectral reflectances  of the different facets of the surface at a single wavelength. A surface reflectance is necessarily superior or equal to $0$ and inferior or equal to $1$ in case of non-florescent materials. Thus, the sum of each of its column is inferior or equal to one. However, in order to satisfy the second condition, an assumption that the surface reflectance is strictly superior to zero should be made. This assumption is met in the spectral reflectance of real surfaces.  $\mathbf{K}$ is also a square matrix, each of its columns represents the geometrical relation between a facet and all the other facets in the surface.  In radiometric terms, this geometrical relation is related to the geometrical extent \footnote{The geometrical extent can be also found under the name étendue}, and when integrated over the hemisphere around an infinitesimal point, this term is equal to $\pi$. Each column of $\mathbf{K}$ represents an integration over a part of the hemisphere, and every term in this matrix is divided by $\pi$, then the sum of each column of this matrix is inferior or equal to one.

Then, both $\mathbf{R}$ and $\mathbf{K}$ are \textit{substochastic} matrices, and:

\begin{equation}
\lim_{n\to\infty} (\mathbf{K}  \mathbf{R})^n = 0.
\end{equation}

\bibliography{refs1}


\end{document}